\documentclass[journal]{IEEEtran}
\IEEEoverridecommandlockouts
\usepackage{cite}
\usepackage{amsmath,amssymb,amsfonts}
\usepackage{algorithmic}
\usepackage{graphicx}
\usepackage{textcomp}
\usepackage{xcolor}
\def\BibTeX{{\rm B\kern-.05em{\sc i\kern-.025em b}\kern-.08em
    T\kern-.1667em\lower.7ex\hbox{E}\kern-.125emX}}
    
\def\BibTeX{{\rm B\kern-.05em{\sc i\kern-.025em b}\kern-.08em T\kern-.1667em\lower.7ex\hbox{E}\kern-.125emX}}

\begin{document}

\title{Tree-based Intelligent Intrusion Detection System \\ in Internet of Vehicles
}

\author{\IEEEauthorblockN{Li Yang$^*$, Abdallah Moubayed$^*$, Ismail Hamieh$^\dagger$, Abdallah Shami$^*$\\}
\IEEEauthorblockA{
$^*$Western University, London, Ontario, Canada \\
e-mails: \{lyang339, amoubaye, abdallah.shami\}@uwo.ca\\}
$^\dagger$National Research Council (NRC), London, Ontario, Canada\\
e-mail: Ismail.Hamieh@nrc-cnrc.gc.ca
}

\maketitle

\begin{abstract}
The use of autonomous vehicles (AVs) is a promising technology in Intelligent Transportation Systems (ITSs) to improve safety and driving efficiency. Vehicle-to-everything (V2X) technology enables communication among vehicles and other infrastructures. However, AVs and Internet of Vehicles (IoV) are vulnerable to different types of cyber-attacks such as denial of service, spoofing, and sniffing attacks. In this paper, an intelligent intrusion detection system (IDS) is proposed based on tree-structure machine learning models. The results from the implementation of the proposed intrusion detection system on standard data sets indicate that the system has the ability to identify various cyber-attacks in the AV networks. Furthermore, the proposed ensemble learning and feature selection approaches enable the proposed system to achieve high detection rate and low computational cost simultaneously.
\end{abstract}


\markboth{Published in IEEE GLOBECOM 2019}
{}

\section{Introduction}
With more vehicles, devices, and infrastructures involved, the conventional vehicular ad hoc networks (VANETs) are gradually evolving into the Internet of Vehicles (IoV) \cite{VANET}. In intelligent transportation systems (ITSs), VANETs enable wireless communications between vehicles and devices, then transform the vehicles and devices into wireless routers or mobile nodes \cite{IOV}. Autonomous vehicles (AV) technology is a fast-paced technology which provides an ideal solution to reduce traffic collisions and related costs. Vehicle-to-everything (V2X) \cite{IOV} technology is at the core of AV, consisting of Vehicle-to-Vehicle (V2V), Vehicle-to-Pedestrian (V2P), Vehicle-to-Infrastructure (V2I), and Vehicle-to-Network (V2N) technologies to enable local and wide area cellular network communications between vehicles, pedestrians, and infrastructures. 

 V2X technology, with the means of wireless communications, aims to involve more Internet of things (IoT) devices. However, some of these devices lack security mechanisms such as firewalls and gateways \cite{intelligent}. AVs are prone to network threats with severe consequences because attacking or maliciously controlling the vehicles on the road poses serious threat to human lives. The potential networking threats include the following attacks. A common type of attack is Denial of services (DoS) attacks launched by sending a large number of irrelevant messages or requests to occupy the node and take control of the resources of the vehicle \cite{attacks}. The attackers perform spoofing attacks such as GPS spoofing to masquerade as legitimate users and provide the nodes with fake GPS information \cite{attacks}. Sniffing attacks such as port scan attacks is another type of attack, which is launched to obtain confidential or sensitive data of the vehicles systems and users \cite{s3d2}. In addition, brute-force attacks are launched by attackers to crack passwords in the vehicle networks or systems \cite{s3d2}. To intrude into the web interface of vehicles or servers, web attacks including SQL injection attacks and cross-site scripting (XSS) can also be implemented by the attackers \cite{s3d2}.

Apart from external communication threats, AVs are also prone to intra-vehicle communication attacks. Controller Area Network (CAN) \cite{GIDS} is a bus communication protocol which enables in-vehicle communications between all Electronic Control Units (ECUs). It provides an efficient error detection mechanism for stable transmission and can reduce wiring cost, weight, and complexity \cite{GIDS}. However, all the ECUs communicate with each other through the CAN bus, which makes the ECUs vulnerable to various attacks if the CAN bus is compromised. In CAN bus communications, attackers can inject malicious messages to monitor the network traffic or launch other hostile attacks and the nodes will inevitably deal with the messages without validating their origins. The message injection attacks on CAN bus can be classified by their aims. Similar to external networks, DoS and spoofing attacks can also be launched on the CAN bus to occupy the resources or provide malicious information such as gear and RPM (revolutions per minute) information. In addition, fuzzy attacks are another common type of attack launched on CAN bus by injecting arbitrary messages to cause the vehicles to show unintended states or malfunction \cite{OTIDS}.

All the above vulnerabilities and threats call for a robust protection system that can repel possible attacks that pose threats to intra-vehicle and external communications in AV systems. Intrusion detection systems (IDSs) are an effective security mechanism to identify the abnormal information and attacks through the network traffic data during the communication between vehicles and other devices. Intrusion detection is often considered as a classification problem; machine learning (ML) methods have been widely used to develop IDSs \cite{data1}. An intelligent IDS\footnote{
code is available at: https://github.com/Western-OC2-Lab/Intrusion-Detection-System-Using-Machine-Learning} is proposed in this paper for network attack detection that can be applied to not only Controller Area Network (CAN) bus of AVs but also on general IoVs. The proposed IDS utilizes tree-based ML algorithms including decision tree (DT), random forest (RF), extra trees (ET), and Extreme Gradient Boosting (XGBoost). A qualified IDS needs to not only achieve a high detection rate, but also have a low computational cost. Therefore, an ensemble learning model, namely stacking, is used to improve accuracy and the feature selection methods are also implemented to reduce computational time. The performance of the proposed IDS is evaluated using multiple standard open-source data sets with results showing high accuracy in detecting intrusions.

This paper makes the following contributions: 
\begin{itemize}
\item Surveys the vulnerabilities and potential attacks in CAN and AV networks. 
\item Proposes an intelligent IDS for both AV and general networks by using the tree structure ML and ensemble learning methods. 
\item Presents a comprehensive framework to prepare network traffic data for the purpose of IDS development. 
\item Proposes an averaging feature selection method using tree structure ML models to improve the efficiency of the proposed IDS and to perform an analysis of network attributes and attacks for network monitoring uses. 
\end{itemize}

This paper is organized as follows: Section II presents the system overview of the proposed IDS and its architecture. Section III discusses the proposed IDS framework in detail.  Section IV provides the results, performance comparison and feature analysis. Finally, Section V concludes the paper.

\section{System Design}
\subsection{Problem statement}

Due to the fact that the AV systems are vulnerable to many types of network threats through different communication mediums, a comprehensive IDS system is proposed to be implemented for both intra-vehicle and external communication networks. This is done to better protect not only the vehicle components but also all the IoT devices involved in the entire IoV. 

The proposed IDS should be able to detect various common intrusions launched on CAN bus and external networks. The main attacks to be considered in this work include DoS attacks on both the intra-vehicle and external communication networks, fuzzy attacks and spoofing attacks on the CAN-bus, sniffing, brute force and web attacks launched on the external networks. The IDS should have a high detection rate to effectively identify most of the attacks and low computational time to improve efficiency.

\subsection{IDS system overview and architecture}
\begin{figure}
     \centering
     \includegraphics[width=8.5cm]{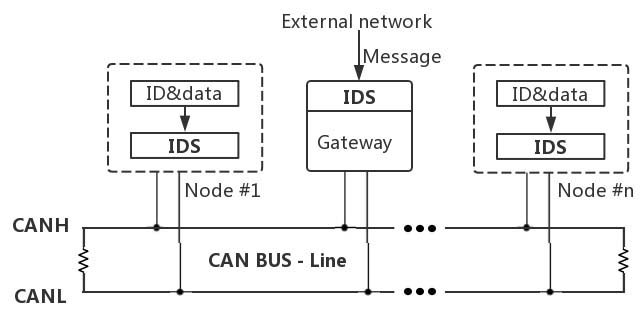}
     \caption{The proposed IDS-protected AV architecture} \label{im}
\end{figure}

To provide protection for both the intra-vehicle and external communications, the proposed IDS is implemented in multiple locations within the AV system. To detect threats on the CAN bus and secure it, the IDS can be placed on the top of the CAN bus to process every transmitted message and ensure the nodes are not compromised \cite{CANim}. Alternatively, the proposed IDS can be placed inside the gateway to secure the external communication networks \cite{exim}. The topology of IDS implementation on vehicle systems is shown in Fig. \ref{im}. 

On the CAN bus, all the nodes are linked, and when a node receives a message mainly consisting of an ID and data field sent by another device connected to the CAN bus, the message is passed through the IDS to check when the signal line of the CAN bus changes from CANH to CANL. Similarly, when a message is transmitted from external network to intranet, it is passed through the IDS inside the gateway and checked.
\begin{figure}
     \centering
     \includegraphics[width=6.5cm]{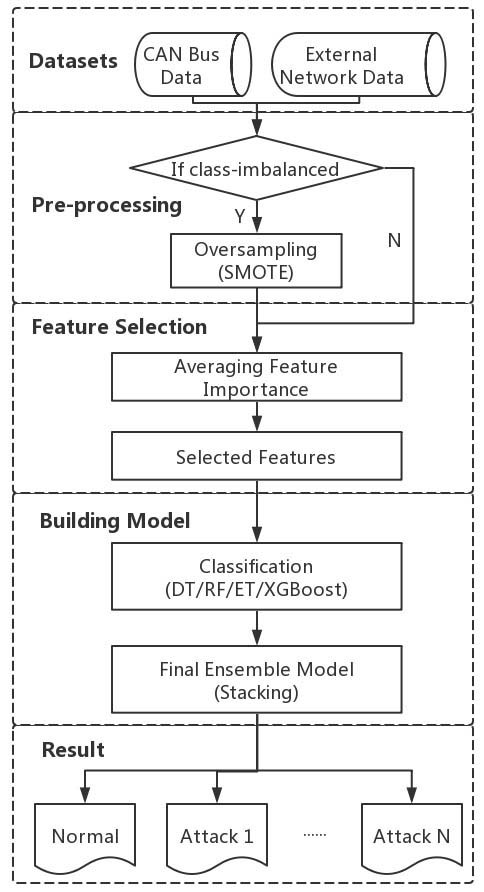}
     \caption{The proposed IDS framework} \label{IDS}
\end{figure}

 To design the IDS, Fig. \ref{IDS} provides an overview of the tree-based ML classification framework for intrusion detection. The process of the proposed model is as follows. Firstly, sufficient network traffic data is collected. Secondly, if the classes of the data set are imbalanced, oversampling is implemented to reduce its impact. At the next stage, feature selection based on averaging feature importance is done to reduce computational cost. After that, four base-models are built to be the input of the stacking ensemble model. At the end, the final model is built to classify the data. 

\section{Proposed IDS Framework}
\subsection{Data pre-processing}

The first step to develop an IDS is to collect sufficient amount of network traffic data under both the normal state and the abnormal state caused by different types of attacks. The data can be collected by the packet sniffers, but they should have suitable network attributes, or named network features, for the purpose of IDS development.  

Firstly, to design an IDS for CAN bus intrusion, since the main CAN bus threats are message injection attacks, the data of CAN messages/frames should be collected and the main features correlated to attacks are CAN IDs and the data field of frames \cite{OTIDS}. 

Regarding the external networks, since they belong to general networks and are prone to various regular network threats, the data with more network attributes should be collected to develop an effective IDS that can detect various types of cyber-attacks. Most of the regular network attributes such as packet length, data transfer rate, throughput, inter-arrival time, flags of TCP and their counts, segment size, and active/idle time should be considered \cite{s3d2}. However, the computational complexity of the proposed IDS may increase dramatically due to the high data dimensionality. Thus, further feature analysis should be done for the external network data.

The collected network data would be pre-processed after a few steps to be better suited for IDS development purpose. Firstly, the data can be encoded with one-hot-vector because it has a certain threshold to help separate normal data and anomalies \cite{GIDS}. On the other hand, ML training is often more efficient with normalized data \cite{intelligent}. Thus, each feature with numerical values is set to the range of 0.0 to 1.0, and each value after normalization $x_{n}$ is indicated as:
\begin{equation}
x_{n}=\frac{x-min}{max-min},
\end{equation}%
where $x$ is the original value, $max$ and $min$ are the maximum and minimum values of each feature.

In addition, network data are often class imbalanced because in real life, networks maintain the normal state most of the time and the attack-label instances are often not enough. To overcome the issue of class-imbalanced data which often results in a low anomaly detection rate, random oversampling and Synthetic Minority Oversampling Technique (SMOTE) \cite{SMOTE} can be used to generate more data in the minority classes that do not have enough data. The basic strategy of random oversampling is to simply make multiple copies of the samples to increase samples in the minority classes. However, the random oversampling method can easily result in over-fitting because the information learned would be very specific instead of generic. On the other hand, the SMOTE algorithm analyzes the minority classes and generates new samples based on them by using the idea of K nearest neighbors. Therefore, SMOTE can generate high-quality samples and is used for the minority classes in the proposed system. 
\subsection{The proposed ML approaches}
In the proposed system, to detect various cyber-attacks, developing the IDS can be considered as a multi-classification problem, and machine learning algorithms are widely used to solve such classification problems \cite{DNS}\cite{ML}. The selected ML algorithms are based on tree structure, including decision tree, random forest, extra trees, and XGBoost.

Decision tree (DT) \cite{DT} is a common classification method based on divide and conquer strategy. A DT comprises decision nodes and leaf nodes, and they represent a decision test over one of the features, and the result class, respectively. Random forest (RF) \cite{RF} is an ensemble learning classifier based on the majority voting rule that the class with the highest votes by decision trees is selected to be the classification result. Similarly, extra trees (ET) \cite{ET} is another ensemble model based on a collection of randomized decision trees generated by processing different subsets of data set. In contrast, XGBoost \cite{XG} is a ensemble learning algorithm designed for speed and performance improvement by using the gradient descent method to combine many decision trees.

Apart from the proposed algorithms, other models such as K-nearest neighbor (KNN) and support vector machine (SVM) \cite{SVM} are also common for classification problems. For the purpose of model selection, the computational complexity of common supervised ML algorithms is calculated. Assuming the number of training instances is $N$, the number of features is $P$, and the number of trees is $T$, we have the following approximations. The complexity of DT is $O(N^2P)$ while the complexity of RF is $O(N^2\sqrt{P}T)$. In addition, ET and XGBoost have a similar complexity of $O(NPT)$. On the other hand, KNN's complexity is $O(NP)$ and the SVM algorithm’s complexity is $O(N^2P)$ \cite{DNS}\cite{time}. However, unlike KNN and SVM, the proposed four tree-based models, DT, RF, ET, and XGBoost, enable multi-threading to save training time. Assuming the maximum number of participating threads of a computer is $M$, the time complexity of DT, RF, ET, and XGBoost reduces to $O(\frac{N^2P}{M})$, $O(\frac{N^2\sqrt{P}T}{M})$, $O(\frac{NPT}{M})$, and $O(\frac{NPT}{M})$, respectively. Therefore, although the original time complexity of the considered algorithms are similar, the four tree structure algorithms have lower computational time due to multi-threading, which is an important reason for choosing these four algorithms.

The other reasons for choosing these algorithms are as follows: 1) Most of the tree structure ML models are using ensemble learning so they often show better performance than other single models such as KNN. 2) They have the ability to handle non-linear and high dimensional data that the proposed network data belongs to. 3) The feature importance calculations are done during the building process of those models, which is beneficial when performing feature selection. 

It is noteworthy that there are some hyper-parameters of the proposed algorithms that need be tuned to achieve better performance. For the DT algorithm, the split measure function is set to be Gini impurity, and the classification and regression trees (CART) model is then built, which shows better performance than using information gain theory to build an ID3 tree. Assuming that $S$ denotes the set of all sub-trees, CART selects the tree in $S$ that minimizes \cite{DT}:

\begin{equation}
C(S) = {\hat L}_n(S)+{\alpha}|S|,
\end{equation}
where $|S|$ is the cardinality of the tree, ${\alpha}$ is a constant and ${\hat L}_n(S)$ is the empirical risk using the tree $S$. Since the deeper tree has more sub-trees, tree depth $D$ is an important parameter of the CART algorithm. 

For RF and ET, since their results are based on the majority voting of many decision trees, the number of decision trees $T$ is another important parameter affecting the performance. $T$ could also be tuned in XGBoost which is also based on the ensemble of numerous trees. Specifically, XGBoost minimizes the  following regularized objective function \cite{XG}:
\\
\begin{equation}
Obj = -\frac{1}{2}\sum_{j=1}^t\frac{G_{j}^2}{H_{j}+\lambda}+\gamma t,
\end{equation}
\\
where $G$ and $H$ represent the sums of the first and second order gradient statistics of the loss function, $t$ is the number of leaves in a decision tree, $\lambda$ and $\gamma$ are the penalty coefficients. Since the gradient statistics are based on the sum of the predicted scores of the $T$ trees, and the number of leaves increases with the increase of the tree depth $D$, $T$ and $D$ have direct impacts on the objective function value of XGBoost.

If the parameters $T$ and $D$ are too small, it leads to under-fitting and if they are too large, it causes over-fitting and additional computational costs. 

The grid search method \cite{grid} is utilized to find the optimal values. To be precise, the training of the models starts from a small number of trees and a short depth. Then, these two values are slowly increased with accuracy evaluated until over-fitting, which is indicated by the dropping of accuracy. Finally, the tree depth $D$ is tuned to 8 and the number of trees $T$ is set to 200. Similarly, the minimum sample split and minimum sample leaf are both tuned from 1 to 10, and finally set to 8 and 3, respectively. Note that the parameter tuning process can be further improved by using other optimization techniques. 

\subsection{Ensemble learning and feature selection}

For the purpose of further accuracy improvement, an ensemble learning technique, stacking, is implemented. Stacking \cite{stacking} is a common ensemble method consisting of two layers where the first layer contains a few trained base predictors, and their output serves as the input of a meta-learner in the second layer to build a strong classifier. The four trained tree structure algorithms serve as the base models in the first layer of the stacking ensemble method, and the singular algorithm with highest accuracy among the four base models is selected to be the meta-classifier in the second layer.

To improve the confidence of the selected features, an ensemble feature selection (FS) technique is utilized by calculating the average of feature importance lists generated by the four selected tree-based ML models. They are chosen for feature selection because tree-based algorithms calculate the importance of each feature based on each single tree, and then average the output of the trees to make the result more reliable. Additionally, different traditional feature selection methods such as information gain, entropy, and Gini coefficient are utilized in the system by setting different parameters in tree-based methods to generate the convincing feature importance. The sum of the total feature importance is 1.0. To select features, the features are ranked with their importance and each feature is added into the feature list from high importance to low importance till the sum of importance reaches 0.9. The other features with the sum of importance less than 0.1 will be discarded to reduce the computational costs.

\subsection{Validation metrics}
Each considered data set is split into five subsets and 5-fold cross-validation is implemented to evaluate the proposed models. Accuracy (Acc), detection rate (DR)/recall, false alarm rate (FAR), and F1 score are the main metrics used to evaluate the proposed method, with their formulas proposed in \cite{saloreview}. The accuracy is the proportion of correctly classified data. However, the data sets may exhibit class imbalance, resulting in a high accuracy of normal data classification but a low attack detection rate. Thus, the detection rate, the ratio between the detected attack data and the total abnormal data, is also calculated for evaluation. A qualified IDS should have a high DR to ensure most of the attacks can be detected and a low FAR to confirm the system does not misreport data for higher DR \cite{RF}. In addition, the F1 score considers both the precision and recall by calculating their harmonic average, which can be used to evaluate the overall performance of the methods. Also, the execution time, mainly the model training time, is used to provide insights into the computational performance.
\section{Performance Evaluation}
\subsection{Datasets description}
To evaluate the proposed IDS, this work considers two different datasets for the purpose of its implementation in both intra-vehicle and external networks as described in Section III-A. The first data set is called "Car-Hacking Dataset" or "CAN-intrusion dataset" which was proposed in 2018 for the purpose of IDS development on CAN bus \cite{GIDS}. On the other hand, to build a comprehensive IDS that can also be effective in external communication networks, a standard IDS data set containing the most updated attack scenarios, named "CICIDS2017" \cite{s3d2}, is considered in this work. The features of the two selected data sets meet our requirements proposed in Section III-A. To prepare better datasets for the IDS development, minor tasks including data combination, missing value removal and new label assignments were done for both datasets based on the methods proposed in \cite{IDS2017}. The specifics of the improved datasets are shown in Table \ref{dc} and \ref{di}.
\subsection{IDS performance analysis}
The proposed system was implemented using Python 3.5 and the experiments were carried out on a machine with 6 Core i7-8700 processor and 16 GB of memory. 

\begin{table}[tbp] \centering%
\caption{Data Type and Size of The CAN-intrusion Dataset}%
\begin{tabular}{|c|c|c|c|}
\hline
\textbf{Class label} & \multicolumn{1}{p{1.5cm}|}{\textbf{Number of instances}} & 
\textbf{Class label} & \multicolumn{1}{p{1.5cm}|}{\textbf{Number of instances}} \\ \hline
Normal & 14,037,293 & RPM spoofing & 654,897\\ \hline
DoS   & 587,521 & Gear spoofing & 597,252  \\ \hline
Fuzzy & 491,847 &- &- \\ \hline
\end{tabular}%
\label{dc}%
\end{table}%

\begin{table}[tbp] \centering%
\caption{Data Type and Size of The CICIDS2017 Dataset}%
\begin{tabular}{|c|c|c|c|}
\hline
\textbf{Class label} & \multicolumn{1}{p{1.5cm}|}{\textbf{Number of instances}} & 
\textbf{Class label} & \multicolumn{1}{p{1.5cm}|}{\textbf{Number of instances}} \\ \hline
    BENIGN & 2,273,097 & Web-Attack & 2,180  \\ \hline
    DoS   & 380,699 & Botnet & 1,966  \\ \hline
    Port-Scan & 158,930 & Infiltration & 36  \\ \hline
    Brute-Force & 13,835 & - & - \\ \hline
\end{tabular}%
\label{di}%
\end{table}%
\begin{table}[tbp] \centering%
\caption{Performance Evaluation of IDS on CAN-intrusion Dataset}%
\begin{tabular}{|c|c|c|c|c|c|}
\hline
    \textbf{Method} & \multicolumn{1}{p{3em}|}{\textbf{Acc (\%)}} & \multicolumn{1}{p{3em}|}{\textbf{DR (\%)}} & \multicolumn{1}{p{3em}|}{\textbf{FAR (\%)}} & \multicolumn{1}{p{3.3em}|}{\textbf{F1 Score}} & \multicolumn{1}{p{4em}|}{\textbf{Execution Time (S)}} \\
    \hline
    KNN \cite{data1} & 97.4  & 96.3  & 5.3   & 0.934 & 911.6 \\
    \hline
    SVM \cite{data1} & 96.5  & 95.7  & 4.8   & 0.933 & 13765.6 \\
    \hline
    DT    & 99.99 & 99.99 & 0.006 & 0.999 & 328 \\
    \hline
    RF    & 99.99 & 99.99 & 0.0003 & 0.999 & 506.8 \\
    \hline
    ET    & 99.99 & 99.99 & 0.0005 & 0.999 & 216.3 \\
    \hline
    XGBoost & 99.98 & 99.98 & 0.012 & 0.999 & 3499.1 \\
    \hline
    Stacking & 100   & 100   & 0.0     & 1.0     & 1237.1 \\
    \hline
    FS RF & 99.99 & 99.99 & 0.0013 & 0.999 & 99.6 \\
    \hline
    FS Stacking & 99.99 & 99.99 & 0.0006 & 0.999 & 325.6 \\
    \hline
\end{tabular}%
\label{pc}%
\end{table}%
\begin{table}[tbp] \centering%
\caption{Performance Evaluation of IDS on CICIDS2017 Dataset}%
\begin{tabular}{|c|c|c|c|c|c|}
\hline
    \textbf{Method} & \multicolumn{1}{p{3em}|}{\textbf{Acc (\%)}} & \multicolumn{1}{p{3em}|}{\textbf{DR (\%)}} & \multicolumn{1}{p{3em}|}{\textbf{FAR (\%)}} & \multicolumn{1}{p{3.3em}|}{\textbf{F1 Score}} & \multicolumn{1}{p{4em}|}{\textbf{Execution Time (S)}} \\
    \hline
    KNN \cite{s3d2} & 96.6  & 96.4  & 5.6   & 0.966 & 9243.6 \\
    \hline
    SVM \cite{SVM2} & 98.01 & 97.58 & 1.48  & 0.978 & 49953.1 \\
    \hline
    DT    & 99.72 & 99.3  & 0.029 & 0.998 & 126.7 \\
    \hline
    RF    & 98.37 & 98.29 & 0.039 & 0.983 & 2421.6 \\
    \hline
    ET    & 93.43 & 93.35 & 0.001 & 0.934 & 2349.6 \\
    \hline
    XGBoost & 99.78 & 99.76 & 0.069 & 0.997 & 1637.2 \\
    \hline
    Stacking & 99.86 & 99.8  & 0.012 & 0.998 & 4519.3 \\
    \hline
    FS XGBoost & 99.7  & 99.55 & 0.077 & 0.996 & 995.9 \\
    \hline
    FS Stacking & 99.82 & 99.75 & 0.011 & 0.997 & 2774.8 \\
    \hline
\end{tabular}%
\label{pi}%
\end{table}%

The results of testing different algorithms on CAN-intrusion data set and CICIDS2017 data set are shown in Tables \ref{pc} and \ref{pi}, respectively. According to Table \ref{pc}, when testing on the CAN-intrusion data set, the tree-based algorithms including DT, RF, ET, and XGBoost used in the proposed system are 2.5\% more accurate than KNN and 3.4\% more accurate than SVM used in \cite{data1}. In addition, multi-threading is enabled in DT, RF, ET, and XGBoost, resulting in a lower execution time compared with KNN and SVM. Since XGBoost has the lowest accuracy and longest execution time among the four tree-based ML models, only the other three algorithms, DT, RF, and ET were selected into the stacking ensemble model, and the single model with best performance, RF, was selected to be the meta-classifier in the second layer. After using stacking to combine the three tree-based models, the accuracy, detection rate and F1 score reaches 100\%, which means that all the trained attacks can be detected. 

Similarly in CICIDS2017 data set, as shown in Table \ref{pi}, most of the used tree-based algorithms shows 1.8-3.2\% higher accuracy, detection rate and F1 score than KNN \cite{s3d2} and SVM \cite{SVM2} except for ET. Hence, only DT, RF, and XGBoost were chosen in the proposed stacking method, and XGBoost was selected to be the meta-classifier of the stacking model. Although the execution time of stacking is longer than any singular tree-based models, it reaches the highest accuracy among the models (99.86\%).

After training the data sets with all the features to obtain the highest performance, the feature selection method proposed in Section III-C was implemented to reduce execution time. From Tables \ref{pc} and \ref{pi}, it can be seen that RF and XGBoost are the singular models with best accuracy on CAN-intrusion data set and CICIDS2017 data set, respectively. As single base models have lower execution time than the ensemble model, these two singular models were also tested after feature selection. For CAN-intrusion data set, the top-4 important features, ‘CAN ID’, 'DATA[5]', 'DATA[3]', 'DATA[1]' were selected for the tests. The results of RF and stacking on CAN-intrusion data set after feature selection, named "FS RF" and "FS Stacking", can be seen from Table \ref{pc}. It shows that the RF and stacking models saved 80.3\% and 73.7\% of the execution time after feature selection, respectively, while still maintaining a high accuracy (99.99\%). For CICIDS2017 data set, 36 out of 78 features were selected and the accuracy of XGBoost and stacking models only decreased by 0.08\% and 0.04\% while saving 39.2\% and 38.6\% of the execution time, respectively, as shown in Table \ref{pi}. Therefore, the tree-based averaging feature selection method enables the IDS to save execution time while maintaining high accuracy.

\subsection{Feature analysis}

\begin{table}[htbp]
  \centering
  \caption{Top-3 Feature Importance by Each Attack}
    \begin{tabular}{|p{4.215em}|p{12em}|p{4.215em}|}
    \hline
    \multicolumn{1}{|c|}{\textbf{Label}} & \multicolumn{1}{|c|}{\textbf{Feature}} & \multicolumn{1}{|p{4.215em}|}{\textbf{Weight}} \\
    \hline
     & Bwd Packet Length Std & 0.1723 \\
\cline{2-3} 
\multicolumn{1}{|c|}{DoS} & Average Packet Size & 0.1211 \\
\cline{2-3}           & Destination Port & 0.0785 \\
    \hline
     & Total Length of Fwd Packets & 0.3020 \\
\cline{2-3}
\multicolumn{1}{|c|}{Port-Scan} & Average Packet Size & 0.1045 \\
\cline{2-3}          & PSH Flag Count & 0.1019 \\
    \hline
     & Destination Port & 0.3728 \\
\cline{2-3}
\multicolumn{1}{|c|}{Brute-Force}          & Fwd Packet Length Min & 0.1022 \\
\cline{2-3}         & Packet Length Variance & 0.0859 \\
    \hline
     & Init\_Win\_bytes\_backward & 0.2643 \\
\cline{2-3}
\multicolumn{1}{|c|}{Web-Attack}          & Average Packet Size & 0.1650 \\
\cline{2-3}          & Destination Port & 0.0616 \\
    \hline
     & Destination Port & 0.2364 \\
\cline{2-3}
\multicolumn{1}{|c|}{Botnet}          & Bwd Packet Length Mean & 0.1240 \\
\cline{2-3}          & Avg Bwd Segment Size & 0.1104 \\
    \hline
     & Total Length of Fwd Packets & 0.2298 \\
\cline{2-3}
\multicolumn{1}{|c|}{Infiltration}          & Subflow Fwd Bytes & 0.1345 \\
\cline{2-3}          & Destination Port & 0.1149 \\
\hline 
    \end{tabular}%
  \label{fi}%
\end{table}%

In order to analyze the features, the proposed feature selection method was tested on the subsets of each attack. The list of top-3 important features and their corresponding weights of each attack is shown in Table \ref{fi}. As Table \ref{fi} shows, the destination port can reflect a DoS, brute force, web attack, and botnet attack. The size of packets is another important feature. For example, the average packet size indicates a DoS attack, port scan attack, and web attack. The packet length in the forward direction is related to port scan, brute-force, and infiltration attacks, while the packet length in the backward direction reflects a DoS, web attack, and botnet. In addition, the variance of the length of the packets in both forward and backward directions reflects the brute force attack, and the count of pushing flags indicates port scan attack. After getting the feature importance list of each attack, the IDSs with other aims like designing a dedicated system for detecting a single type of attack can be developed by selecting the important features based on the list. On the other hand, the important features can be selected as key attributes to be monitored by network supervisors. If those attributes change abnormally, the attacks can be detected quickly.

\section{Conclusion}
As autonomous vehicles and connected vehicles are vulnerable to various cyber-attacks, IDS is one of the efficient solutions to detect the launched network attacks and secure the vehicle networks. In this paper, we presented an IDS based on tree-based machine learning algorithms to detect threats both on CAN bus and external networks. The SMOTE oversampling method and tree-based averaging feature selection approaches were also introduced to reduce the impact of class imbalance and computational cost. To evaluate the proposed IDS, it was tested on two data sets for both intra-vehicle and external networks. The results on both data sets show that the proposed system has 2-3\% higher accuracy, detection rate, F1 score and lower false alarm rate than the methods proposed in the literature. Moreover, unlike other proposed methods, our work combines all the subsets of the data sets and develops an IDS that can detect various attacks instead of only single type of attack on each run. The accuracy of CAN intrusion and CICIDS2017 data set reaches 100\% and 99. 86\%, while the computational time was reduced by 73.7\% to 325.6s and by 38.6\% to 2774.8s, respectively. In future work, the results of the proposed IDS on CICIDS2017 data set can be further improved by using optimization techniques such as particle swarm optimization and Baysian optimization to tune the hyper-parameters of the proposed algorithms.

\end{document}